# E Pluribus Unum Interpretable Convolutional Neural Networks

George Dimas[1], Eirini Cholopoulou[1] and Dimitris K. Iakovidis[1]✉

[1] Dept. of Computer Science and Biomedical Informatics, School of Science, University of Thessaly, Lamia, Greece

The adoption of Convolutional Neural Network (CNN) models in high-stake domains is hindered by their inability to meet society's demand for transparency in decision-making. So far, a growing number of methodologies have emerged for developing CNN models that are interpretable by design. However, such models are not capable of providing interpretations in accordance with human perception, while maintaining competent performance. In this paper, we tackle these challenges with a novel, general framework for instantiating inherently interpretable CNN models, named E Pluribus Unum Interpretable CNN (EPU-CNN). An EPU-CNN model consists of CNN sub-networks, each of which receives a different representation of an input image expressing a perceptual feature, such as color or texture. The output of an EPU-CNN model consists of the classification prediction and its interpretation, in terms of relative contributions of perceptual features in different regions of the input image. EPU-CNN models have been extensively evaluated on various publicly available datasets, as well as a contributed benchmark dataset. Medical datasets are used to demonstrate the applicability of EPU-CNN for risk-sensitive decisions in medicine. The experimental results indicate that EPU-CNN models can achieve a comparable or better classification performance than other CNN architectures while providing humanly perceivable interpretations.

**Keywords:** Interpretable Classification, Convolutional Neural Networks, Generalized Additive Models, Medical Applications

## 1. Introduction

Recently the commercial applicability of Machine Learning (ML) algorithms has been regulated through legislation acts that aim at making the world 'fit for the digital age' with requirements, safeguards, and restrictions regarding ML and automatic decision-making in general [1]. A crucial aspect regarding the compatibility of ML models concerning these regulations is interpretability. But how is the interpretability of ML models defined? According to the recent literature [2], interpretability refers to a passive characteristic of a model, indicating the degree to which a human understands the cause of its decision. Hence, the provided interpretations of the decision-making process of a model can limit its opaqueness and earn users' trust, *e.g.*, by offering interpretations for risk-sensitive decisions in medicine. In real-world tasks, the discriminative power of ML models, as expressed by their performance measures, *e.g.*, their predictive accuracy, is regarded as an insufficient descriptor of their decisions [3].

Various approaches have tackled interpretability from a *post hoc* perspective, *i.e.*, using methods that receive as input a fitted black-box to determine the causality of its predictions [4]. *Post hoc* approaches include image perturbation methods applied on the network by masking, substituting features with zero or random counterfactual instances, occlusion, conditional sampling, *etc.* Such approaches aim at revealing impactful regions in the image that affect the classification result [5, 6]. Other *post hoc* methodologies handle the interpretation problem by constructing simple proxy models, with similar behavior to the original model and implement the perturbation notion at a feature-level [7, 8]. This approach limits the credibility of the explanations, since the proxy model only approximates the computations of the black box [9]. Another set of techniques that reduce the complexity of operations to achieve interpretability utilize the gradient that is backpropagated from the output prediction to the input layer. These methods construct saliency maps by visualizing the gradients to present areas that are considered important by the network [10]; solely relying on their explanations, however, can be misleading [11]. In general, these methods aim at interpreting the inference of a deep learning model after its development and training, which can lead to unreliable interpretations [12].

A different approach to interpretability is the development of ML models that are interpretable by design, *e.g.*, decision trees, lists, and sets [13]. Such models are also referred to as inherently interpretable, and usually, introduce a trade-off between interpretability and accuracy. The structure of such a model is simpler; thus, its predictive performance may be inferior to that of a more complex black-box model. However, this trade-off might be preferable in high-risk decision-making domains due to the importance of understanding, validating and trusting ML models [14]. CNNs with embedded feature guiding and self-attention mechanisms in their architecture, can also be regarded as inherently interpretable [15]. These mechanisms derive interpretations by visualizing saliency maps and CNN features indicating certain concepts on the input image [16]. However, such models usually do not associate the saliency maps with human-perceivable features, and do not account for the contribution of these salient regions to the result. Other methods quantify the alignment of predefined concepts with learned filters in different layers of a network or aim towards the disentanglement of features [17] however, they do not address the direct contribution of the concept representations to the prediction [18]. Also, training such models requires a considerable manual effort for additional annotations with respect to the human-understandable concepts illustrated in each image [19]. Approaches extending regular CNNs to encode object parts in deeper convolutional layers, have also been proposed; nevertheless, they usually result in performance degradation [20]. Another approach is to leverage the intelligibility and expressiveness of Generalized Additive Models (GAMs) [21], which are recognized for their interpretability [22]. The interpretation of a GAM is based on observations associating the effect of each input feature to the predicted output. A variety of applications incorporate GAMs into their methodology to leverage their expressiveness in domains such as healthcare [23]. GAMs based on Multilayer Perceptrons (MLPs) [24], were recently proposed for interpretable data classification and regression; however, these particular models are not tailored for contemporary, CNN-based, computer vision tasks.

State-of-the-art interpretable CNN models usually exploit the information deriving from saliency maps, indicating image regions on which the model focuses its attention; however, it is not apparent how these regions contribute to the predictions. A recent relevant methodology incorporates interpretable components into a CNN model to explain its predictions [25]; nevertheless, the provided interpretations are intertwined with predefined edge kernels, and the selection of the color components does not consider any aspects of human perception. In general, there is a lack of methodologies that could explain the classification of an image based on perceptual features, *i.e.*, features such as color and texture, described in a way that can be easily perceived and interpreted by humans [26].

In this paper, we propose a novel framework for the construction of inherently interpretable CNN models for computer vision tasks, motivated by the need for perceptual interpretation of image classification. The proposed framework is named after the Latin expression *E Pluribus Unum Interpretable CNN* (EPU-CNN), which means "out of many, one" interpretable CNN. A major advantage of the proposed framework is that it is generic, in the sense that it can be used to render conventional CNN models interpretable. Given a base CNN architecture, an EPU-CNN model can be constructed as an ensemble of base CNN sub-networks, by following the GAM approach. The EPU-CNN framework requires that each sub-network of the model receives a set of orthogonal (complementary) perceptual feature representations of the same input image. EPU-CNN is therefore scalable as it can accommodate an arbitrary number of parallel sub-networks corresponding to different perceptual features. The sub-networks are jointly trained and working as one, to automatically generate interpretable class predictions. An EPU-CNN model associates the perceptual features with salient regions, as computed by the different sub-networks, and it explains a classification outcome by indicating the relative contribution of each feature to this outcome.

To the best of our knowledge, EPU-CNN is the first framework based on GAMs for the construction of interpretable CNN ensembles, regardless of the base CNN architecture used and the application domain. Unlike current ensembles, the models constructed by EPU-CNN enable interpretable classification based both on perceptual features and their spatial expression within an image; thus, it enables a more thorough and intuitive interpretation of the classification results. Notably, ensembling shallower CNN architectures can be more efficient than training a single large model [27]. Furthermore, unlike previous interpretable CNN models [20, 28], the classification performance of EPU-CNN models is comparable to or higher than that of their non-interpretable counterpart, which in the case of EPU-CNN is the base CNN model. This is demonstrated with an extensive experimental evaluation on various biomedical datasets, including datasets from gastrointestinal endoscopy and dermatology, as well as a novel contributed benchmark dataset, inspired by relevant research in cognitive science [29].

## 2. Methodology

As a framework, EPU-CNN follows the GAM approach for the construction of interpretable image classification models. GAMs represent a class of models extending linear regression models by using a sum of unknown smooth functions $\sum f_i(x_i)$, $i = 1, 2, \ldots, N$. A GAM is formally expressed as follows:

$$g(\mathbb{E}[Y \mid \boldsymbol{x}]) = \beta + \sum_{i=1}^{N} f_i(x_i) \qquad (1)$$

where $\boldsymbol{x} = (x_1, x_2, \ldots, x_N)^{\mathrm{T}}$, $\boldsymbol{x} \in \mathbb{R}^N$, denotes an input feature vector, $g(\cdot)$ is a link function ($e.g.$, logit), $\beta$ is a bias term and $\mathbb{E}[Y \mid \boldsymbol{x}]$ denotes the expected value of the response variable $Y$, given an input $\boldsymbol{x}$. Each $f_i(\cdot)$, represents a univariate smooth function, $f_i : \mathbb{R} \to \mathbb{R}$, mapping each $x_i \in \mathbb{R}$ to a latent representation, $f_i(x_i)$, through which, $x_i$ participates to the result. This structure is easily interpretable because it enables the user to explore how each input variable $x_i$ affects the predicted output.

The EPU-CNN framework considers Eq. (1) as a template to construct an interpretable ensemble of CNNs from a conventional, non-interpretable, CNN base model (Fig. 1). The sub-networks of the ensemble are arranged in parallel, and each sub-network has the same architecture with the base model. Each sub-network receives a different input, which should be a perceptual feature representation of an input image. This representation will be referred to as *Perceptual Feature Map* (PFM) of an input image, and it can be obtained by an image transformation revealing a physical property of choice that can be easily perceived and interpreted by humans over the input image space, *e.g.*, color and texture [26]. The number of sub-networks is determined by the number of different PFMs required to render a CNN interpretable for a particular application. Considering that each sub-network of an ensemble with a parallel topology should receive inputs with complementary information[30], the PFMs should be orthogonal. Let us consider $N$ different PFMs $I_i$, $i = 1, 2, ..., N$, of an input image $I$. Each $I_i$ is provided as input to a corresponding sub-network $\boldsymbol{C}_i(\ \cdot\ ;\ \eta_i)$, which is parametrized by $\eta_i$, and trained jointly with the rest of the sub-networks. Hence, the input of an EPU-CNN model is a tensor $\boldsymbol{I} = (I_1, I_2, ..., I_N)$ with dimensions of $N{\times}H{\times}W$, where $N$, $H$, and $W$ denote the number, height, and width of the PFMs, respectively. Each sub-network provides a univariate output $\boldsymbol{C}_i(I_i; \eta_i)$.

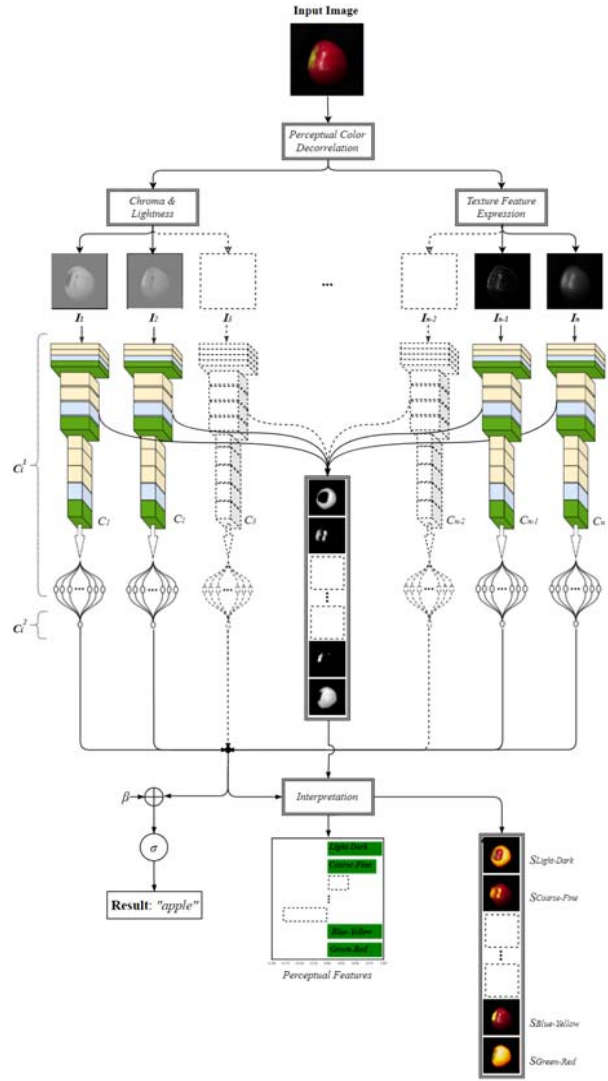

Figure 1. Outline of the EPU-CNN framework.

The output of the EPU-CNN ensemble is computed by summing up all $\boldsymbol{C}_i(I_i; \eta_i)$, $i = 1, 2, ..., N$. The output of each $\boldsymbol{C}_i(I_i; \eta_i)$ can be regarded as a Relative Similarity Score (RSS), quantifying the resemblance of image $I$ to a class with respect to $I_i$. Considering a binary classification problem, RSS takes values within the range of [-1, 1]. It represents the degree of similarity of an input image to a particular class, with respect to a particular PFM $I_i$. An absolute RSS value closer to 1 implies a greater similarity, whereas a positive or negative sign of the RSS associates the similarity with the one class or the other. By visualizing these scores, it becomes easier for a human to understand how each $I_i$ affects a classification result of the EPU-CNN model. Furthermore, by examining the layer activations of $\boldsymbol{C}_i(I_i; \eta_i)$, the scores can be associated with respective image regions; thus, enabling a deeper interpretation of the classification result, based on the spatial arrangement of the observed features within the input image. The details about the PFMs considered in this study, the formulation of the classification model, and its interpretable output, are described in the following paragraphs.

## 2.1 Opponent Perceptual Feature Maps

In this study the generation of PFMs is motivated by the theory of human perception of color vision proposed by Hering in the 1800's, and the opponent-process theory proposed in the 1950's by Hurvich and Jameson [31]. The behavior of a cell in the retina of the human visual system is determined by a pattern of photoreceptors, which comprises a receptive field. Receptive fields have a center-surround organization, which causes the cell to exhibit spatial antagonism, *e.g.*, a cell that is excited by a light stimulus in the center of its receptive field will be inhibited by a light stimulus in the annulus surrounding the excitatory center. There are different types of photoreceptors, with different sensitivities to light frequency and intensity, responding differently to chromatic and luminance variations. Depending on the type of the photoreceptors, receptive fields can be color-opponent or spatially-opponent without being color-opponent [32]. Studies have provided indications that the transmitted stimuli to the retina can be decomposed into independent luminance and chromatic-opponent sources of information, and that the chromatic and luminance information of an image are processed through separate pathways by the human visual system [33, 34]. Also, computer vision experiments have indicated the encoding of the chromatic and luminance components separately, as a more effective approach for image recognition [34, 35].

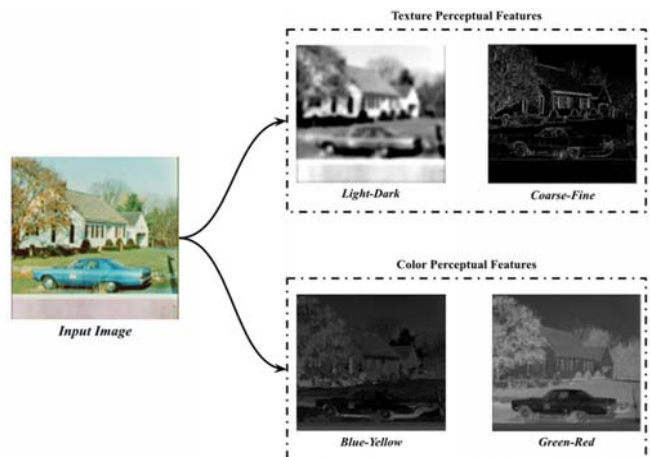

Figure 2. Illustration of the opponent perceptual features utilized by EPU-CNN.

Motivated by these studies, the proposed framework considers an opponent representation of the input images, focusing on both color and texture, which are two decisive properties for image understanding [26]. Also, color and texture provide cues enabling inferences about the shapes of objects and surfaces present in the image. Opponent color spaces have been proposed to cope with drawbacks of the RGB color space, such as the high correlation between the R, G and B color components, and its incompatibility with human perception. Representative examples include, Ohta's color space, which is obtained as a linear transformation of RGB, and it has been proposed in the context of color image segmentation, and CIE-*Lab*, which is obtained as a non-linear transformation of RGB, proposed as a device independent, perceptually uniform color space (*i.e.*, a color space where a given numerical change corresponds to similar perceived change in color) [36]. Considering the effectiveness of CIE-*Lab* in numerous applications in computer vision, especially in biomedicine [37], in this study CIE-*Lab* is considered as a basis for the derivation of three PFMs, corresponding to its components. All the components of CIE-*Lab* are approximately orthogonal. Components *a* and *b* encode two antagonistic colors that cannot be perceived together simultaneously, *e.g.*, there is no "reddish-green" or "bluish-yellow" color. Specifically, component *a*, expresses the antagonism between *green-red* hues (redness is expressed for $a > 0$, and greenness is expressed for $a < 0$), and component *b* expresses the antagonism between *blue-yellow* hues (yellowness is expressed for $b > 0$, and blueness is expressed for $b < 0$). The *L* component represents the perceptual lightness, which expresses an antagonism in luminance, between light and dark. This component, which is practically a greyscale representation of the RGB image, is usually characterized by the highest variance, as it concentrates rich information about the texture of the image contents [34].

In the field of computer vision, several studies have been based on spatial frequency representations of images, to effectively model texture for machine perception [38]. Aiming to the interpretation of the classification outcomes based on perceptual texture characteristics, the *L* component is further analyzed with respect to its spatial frequency. The human eye has a capacity to focus on the right range of spatial frequencies to capture the relevant details of images; thus, visual perception treats images at different levels of resolution. At lower resolutions, these details correspond to larger physical structures in a scene, whereas at higher resolutions the details correspond to smaller structures. The concept of multiresolution image representation can be modeled by the 2D Discrete Wavelet Transform (DWT) [39]. This representation is computed by decomposing the original image using a wavelet orthonormal basis. The computation of the 2D DWT is performed efficiently using the *à trous* algorithm, which is based on convolutions of the image with a pair for low and high-pass filters, called Quadrature Mirror Filters (QMFs), and dyadic down-sampling. A multilevel 2D DWT, focusing on different bands of non-overlapping spatial frequencies, can be performed by successive application of the 1-level 2D DWT on the filtered image with the lowest frequencies. The lowest frequency image of the last level represents a smooth *approximation* of the input image, where the different structures, *e.g.*, objects, parts of objects and background, can be easier segregated upon their intensity. Based on this observation, in the context of EPU-CNN, the approximation image of the third level of the 2D DWT was selected as a PFM representing the *light-dark* antagonism with sufficiently less noise than the original *L* component. The higher frequency bands can be used as PFMs representing image texture at a higher detail. The selection of frequency bands depends on the application context and the pursued interpretation detail. In this study, the highest frequency band of the first level of the 2D-DWT was selected to represent the antagonistic concept of *coarse-fine* texture by exploiting the edges of the image becoming clearer at that level. The concept of coarse-fine texture is associated with the density of image edges per unit area [40], because finer textures tend to have a higher density of edges per unit area than coarser textures. Such edge-based representations are perceptually meaningful for the discrimination of different objects in a scene [41]. Considering that after each level of the 2D DWT the resolution of the image of the previous level is halved, and that the architecture of the base CNN models depends on the dimensions of the input image, the filtered images obtained after the application of the 2D DWT are up-sampled to match the size of the input image *I*. Thus, in this study the input tensor of an EPU-CNN model is formed as $I = (I_1, I_2, I_3, I_4)$, where $I_1$ and $I_2$ are the PFMs corresponding to the *light-dark* and *coarse-fine* concepts respectively, $I_3 = b$ corresponds to *blue-yellow*, and $I_4 = a$ corresponds to the *green-red* concept. An example illustrating the opponent PFMs used in this study, is provided in Fig. 2.

## 2.2 Classification Model

Given an input tensor $I$ composed of $N$ input PFMs, an EPU-CNN model performs feature extraction and classification. An EPU-CNN model is constructed from $N$ CNN sub-networks $C_i$, with each sub-network receiving a PFM $I_i$, $i = 1, 2, …, N$, as input. Each $C_i$ can be regarded as a function $C_i(\cdot\ ; \eta_i)$, $C_i : X^{H \times W} \to Z$, where $X^{H \times W}$ and $Z$ are the input and univariate output space of each $C_i$, respectively. A sub-network $C_i$ consists of two parts: *a)* a feature extractor $C_i^1(\cdot\ ; \omega_i)$ parametrized by $\omega_i$; and *b)* a univariate function $C_i^2(\cdot\ ; \theta_i)$ parametrized by $\theta_i$. Thus Eq. (1) becomes:

$$g(\mathbb{E}[Y \mid I; \theta_{\{N\}}, \omega_{\{N\}}]) = \beta + \sum_{i=1}^{N} C_i^2(C_i^1(I_i; \omega_i); \theta_i) \qquad (2)$$

where $C_i^1$ represents a feature extraction model composed of a CNN followed by a Fully Connected Neural Network (FC-NN), that utilizes activation functions, which are not conditioned to be smooth, and $C_i^2$ represents a single

FC-NN layer utilizing a smooth activation function that provides the final univariate output of a CNN sub-network, $\omega_{\{N\}} = \{\omega_1, \omega_2, \ldots \omega_N\}$ and $\theta_{\{N\}} = \{\theta_1, \theta_2, \ldots \theta_N\}$ are the parameters of $\boldsymbol{C}_i^1, \boldsymbol{C}_i^2$, respectively. Equation (2) encapsulates the properties and definition of GAMs while extending its capacity to exploit CNN models for computer vision tasks. The feature extractor $\boldsymbol{C}_i^1$, can be implemented by a conventional CNN architecture, whereas the number of output neurons and the activation function of $\boldsymbol{C}_i^2$ should be considered so to appropriately represent the classification outcome, in a binary or multiclass setting. In the context of binary classification, which is considered in this study, $\boldsymbol{C}_i^2$ is formulated with a single output neuron and the hyperbolic tangent (*tanh*) activation function, resulting in sub-network responses within the range of [-1, 1]. Ultimately, this allows to intuitively express the contribution of each feature to the final prediction, as positive or negative contribution with respect to a class label. Since EPU-CNN is applied in the context of binary classification, we chose the final output of an EPU-CNN model to be within the interval of [0, 1]. However, the formulation of EPU-CNN presented in Eq. (2) indicates that the final output can fall out of the range of [0, 1], *i.e.*, given a $N$ number of $\boldsymbol{C}_i$ and a bias term $\beta$ the right-hand part of Eq. (2) provides values that fall within the range of $[\beta - N, \ \beta + N]$.

Accordingly, the *logit*($\cdot$) function, defined as:

$$logit(x) = -log\left(\frac{1}{x} - 1\right) \tag{3}$$

can be used as a suitable link function, $g(\cdot)$. Considering that the inverse of *logit* is the *log-sigmoid* function $\sigma$, Eq. (2) can be rewritten as:

$$\mathbb{E}[Y \mid \boldsymbol{I}; \theta_{\{N\}}, \omega_{\{N\}}] = logit^{-1}\left(\beta + \sum_{i=1}^{N} \boldsymbol{C}_i^2(\boldsymbol{C}_i^1(I_i; \omega_i); \theta_i)\right) \tag{4}$$

or

$$EPU_{CNN}\left(\boldsymbol{I}; \eta_{\{N\}}\right) = \sigma\left(\beta + \sum_{i=1}^{N} \boldsymbol{C}_i(I_i; \eta_i)\right) \tag{5}$$

where $\eta_{\{N\}} = \{\eta_1, \eta_2, \ldots, \eta_N\}$. By utilizing the *log-sigmoid* function we bound the output of EPU-CNN within the desirable range of [0, 1] suitable for binary classification applications. Equation (5) is a formal representation of an EPU-CNN model as illustrated in Fig. 1. To train an EPU-CNN model in the context of binary classification, the

Binary Cross Entropy (BCE) is chosen as a loss function to be minimized:

$$EPU_{CNN}: \begin{cases} argmin_{\eta_1}\left(-\frac{1}{k}\sum_{j=1}^{k} y_j log\left(EPU_{CNN}(\mathbf{I}_j; \eta_{(N)})\right) + (1-y_j)log\left(1-EPU_{CNN}(\mathbf{I}_j; \eta_{(N)})\right)\right) \\ \qquad\qquad\qquad\qquad\qquad\qquad . \\ \qquad\qquad\qquad\qquad\qquad\qquad . \\ \qquad\qquad\qquad\qquad\qquad\qquad . \\ argmin_{\eta_N}\left(-\frac{1}{k}\sum_{j=1}^{k} y_j log\left(EPU_{CNN}(\mathbf{I}_j; \eta_{(N)})\right) + (1-y_j)log\left(1-EPU_{CNN}(\mathbf{I}_j; \eta_{(N)})\right)\right) \end{cases} \qquad (6)$$

where $j = 1, 2, 3\ldots, k$, $EPU_{CNN}(\mathbf{I}_j; \eta_{\{N\}})$ is the class probability of $\mathbf{I}_j$ and $y_j$ is the ground truth label of $\mathbf{I}_j$. As it can be observed from Eq.(6), the total error of the EPU-CNN, deriving from the responses of the CNN ensemble consensus, is used to update the parameters of each $\mathbf{C}_i(\cdot; \eta_i) = \mathbf{C}_i^2(\mathbf{C}_i^1(\cdot; \omega_i); \theta_i)$ of the parallel sub-network ensemble topology, simultaneously. It is worth noting that an EPU-CNN model can also be adapted for multiclass datasets, *e.g.*, using $n > 1$ output neurons instead of one, in the case of $n > 1$ classes. Then, the network's output can be interpreted by considering the contribution of the multiclass classification outcome of each CNN sub-network to the final classification result (see Section 3.6).

## 2.3 Interpretable Output

Given an input image, EPU-CNN provides three outputs, as illustrated in Fig. 1, namely, *a)* the predicted class $EPU_{CNN}(\mathbf{I}; \eta_{\{N\}})$; *b)* a set of RSSs $\mathbf{C}_i(I_i; \eta_i)$, $i = 1, 2, \ldots, N$, explaining why the image is classified in that class; and *c)* a set of Perceptual Relevance Maps (*PRMs*) $S_i$ explaining which image regions are responsible for each RSS. Figure 3 illustrates the provided outputs of the model for two images that belong to different classes. The classification result is indicated as a textual label characterizing the input image, and the RSSs are visualized through bar-charts. Each bar-chart consists of horizontal red or green colored bars, indicating the magnitude of resemblance that each $I_i$ is estimated to have for the banana and apple class, respectively. Additionally, the model provides with respect to each $I_i$, areas (PRMs) highlighting their resemblance to the predicted class. The color scaling from orange to yellow regions of the maps indicates the ascending intensity of activation.

Image-specific visualizations of RSSs enable the interpretation of the classification process of unlabeled input images. This is the most important aspect of an EPU-CNN model. For example, the image of Fig. 3(a), is classified as a banana, because all PFMs, *i.e., light-dark*, *coarse-fine*, *blue-yellow* and *green-red*, as indicated by the respective RSSs, guide the prediction towards the banana class, which corresponds to negative $C_i(I_i; \eta_i)$ responses (red). Accordingly, the image of Fig. 3(b), is classified as an apple, because all PFMs guide the prediction towards the apple class, *i.e.*, positive $C_i(I_i; \eta_i)$ responses (green). However, it is not necessary for all the $C_i(I_i; \eta_i)$ responses to be negative for an image to be classified as a banana, since EPU-CNN models consider the consensus of the sub-networks.

Perceptual Relevance Maps, $S_i$, are generated to visually inspect the relevant regions of the input image $I$ with respect to each RSS $C_i(I_i; \eta_i)$. Let $F_i^l = (f_{i,l}^1, f_{i,l}^2, f_{i,l}^3 ..., f_{i,l}^n)$ indicate a tensor of feature maps with $F_i^l \in \mathbb{R}^{n \times h \times w}$, where $n$, $h$, $w$ denote the depth, height and width of $F_i^l$, and $f_{i,l}^n \in \mathbb{R}^{h \times w}$, as computed by a convolutional layer $l$ of a $C_i$. The selection of $l$ is intertwined with its capacity to highlight regions that contribute to the derivation of $C_i(I_i; \eta_i)$. The deeper the layer $l$ that $F_i^l$ is extracted from, the more approximate the correspondence among the feature maps $f_{i,l}^n$ and the input image $I$; thus, a middle layer $l$ of $C_i$ is considered for the construction of each $S_i$ [42] (Section 3.4).

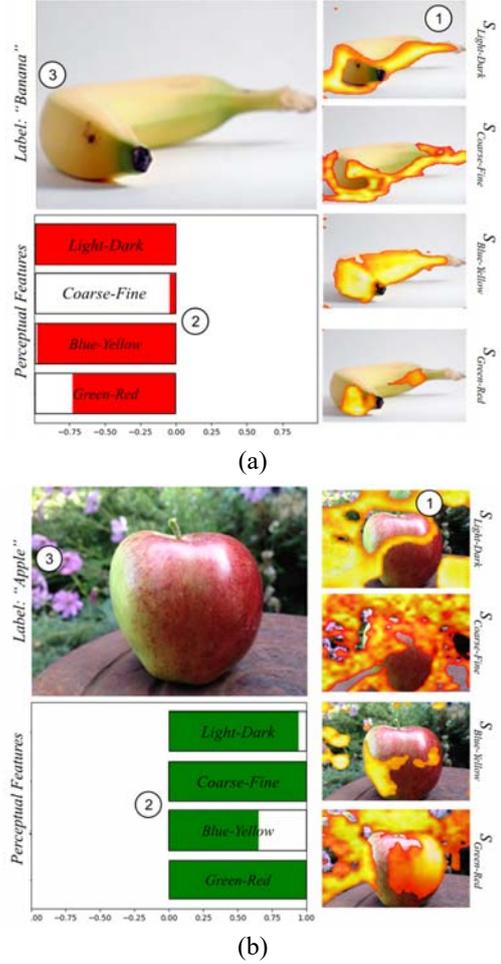

(a)

(b)

Figure 3. Example of EPU-Net output visualization using bar-charts and saliency maps. The numbering indicates the interpretation order of EPU-CNN output. The label field indicates the predicted label. (a) Interpretation of an image classified as a banana. (b) Interpretation of an image classified as an apple.

To quantify the amount of information that each $f_{i,l}^n$ encodes, we compute the Shannon Entropy (SE) scores. Then, half of the most informative $f_{i,l}^n$, *i.e.*, $f_{i,l}^n$ that correspond to the highest entropy scores, are aggregated to construct the $S_i$. The aggregation is performed by averaging $f_{i,l}^n$ features maps which results to the initial $S_i$ estimation. Then $S_i$ is further refined, by applying a thresholding method that maximizes the entropic correlation between the foreground and background of $S_i$, for maximum information transfer [43]. The entropy-based thresholding operation is performed to exclude values associated with lower saliency and communicate to the user the most informative regions. An example of different $S_i$ of input images $I$ can be seen in Fig. 3. The generated $S_i$ illustrated in Fig. 3 are overlaid on the input images (Fig. 3(a, b)). The highlighted regions indicate the spatial association of similarity scores $\boldsymbol{C}_i(I_i; \eta_i)$ with the respective input image. Moreover, the numbers in the images of Fig. 3 indicate in which order the different outputs of an EPU-CNN can be considered by the user. Initially a user can examine the regions that are highlighted by the generated PRMs of each PFM (1). Subsequently, these regions are participating to the classification outcome, towards either class, with a magnitude that is indicated by the RSSs (2). Finally, the PRMs (1) along with the RSSs (2) can assist the user to interpret the class prediction of an EPU-CNN (3).

## 3. Experiments and Results

### 3.1 Datasets

EPU-CNN was trained and evaluated on six different datasets. Initially, a dataset specifically created for the evaluation of the interpretability capabilities of EPU-CNN was considered. The purpose of using this dataset was to demonstrate the capabilities of EPU-CNN with clear, simple, and perceptually meaningful examples. Considering biomedicine as a critical application area for explainable and interpretable artificial intelligence (AI), four well-known biomedical benchmark datasets, consisting of endoscopic and dermoscopic images, was used for further evaluation. Furthermore, to demonstrate the generality of the proposed approach, a well-known benchmark dataset for real image classification was considered.

*Interpretability Dataset*: For the purposes of this study, a novel dataset was constructed, named Banapple. The dataset consists of images of bananas and apples. It was created by collecting images, under the Creative Commons license, from Flickr. The images illustrate bananas and apples with variations regarding the color, placement, size, and background. The motivation for the construction of this dataset stems from studies in cognitive science, where human perception is investigated using examples with discrete properties of bananas and apples[29]. The experiments performed aim to demonstrate that EPU-CNN is capable of capturing the discriminative characteristics of bananas and apples by the perceptual features it incorporates, *i.e.*, apples have a circular shape and usually red color, whereas bananas have a bow-like shape and usually a yellow color. In addition, samples that deviate from the average appearance of these objects can provide insights regarding the reliability of the interpretation of the model.

*Endoscopic Datasets*: Publicly available datasets of endoscopic images were considered for the evaluation process. Namely, KID [44], Kvasir [45] and a dataset that was part of the MICCAI 2015 Endovis challenge [46]. The KID dataset consists of 2,352 annotated wireless capsule endoscopy (WCE) images of abnormal findings *i.e.,* inflammatory, vascular and polypoid lesions as well as images depicting normal tissue from the esophagus, stomach, small bowel and colon. The Kvasir dataset consists of images of the gastrointestinal (GI) tract, annotated and verified by medical experts. These include 4,000 images of anatomical landmarks, *i.e.,* Z-line, pylorus and cecum, and pathological findings of esophagitis, polyps and ulcerative colitis. The dataset also contains sets of images related to endoscopic polyp removal that were not utilized for this work. The MICCAI 2015 Endovis challenge dataset consists of 800 gastroscopic images of normal and abnormal findings, such as gastritis, ulcer, and bleeding.

*Dermoscopic Dataset*: The evaluation process of EPU-CNN has also included the International Skin Image Collaboration Challenge 2019 (ISIC2019) dermoscopic image collection. ISIC2019 challenge provides a publicly available archive of 25,331 dermoscopic images of eight different categories of skin lesions, namely, melanoma, melanocytic nevus, carcinomas (both of basal and squamous cells), actinic and benign keratosis, dermatofibroma, and vascular lesions. These images were used to construct three different binary classification problems: *a)* melanomas *vs.* melanocytic nevi (*Me. vs. Ne.*); *b)* carcinomas *vs.* melanocytic nevi (*Ca. vs. Ne.*) and *c)* carcinomas *vs.* melanomas (*Ca. vs. Me.*). The tasks *a)* and *b)* are characterized as a classification between abnormal (carcinomas, melanomas) and normal (melanocytic nevus) skin lesions whereas task *c)* discriminates two abnormal categories of different incidence and survival rates, *i.e.*, melanomas have higher mortality rates than carcinomas [47]. Task *a)* comprised of 9000 images whereas task *b)* and *c)* 8200 and 8500 images respectively.

*CIFAR-10*: To demonstrate the generality of the proposed framework, EPU-CNN was further validated on the long-standing benchmark dataset CIFAR-10. CIFAR-10 consists of 60,000 color images of natural objects that belong to 10 different classes. The dataset is split in 50,000 training and 10,000 test images and each class comprises 6,000 images with a size of $32 \times 32$.

## 3.2 Classification Performance Assessment

For the comparison of the classification performance of EPU-CNN, we selected three well established CNN models, namely, $VGG_{16}$ [48], $ResNet_{50}$ [49] and DenseNet$_{169}$ [50] and an inherently interpretable CNN model abbreviated as TT [51]. In this study, the $VGG_{16}$ was used as a base for the TT model. The same training parameters, $i.e.$, batch size, optimization algorithm and data augmentation, were applied on all networks involved in the evaluation process. In detail, the batch size was set to 64 and as an optimization algorithm the Stochastic Gradient Decent was used; the training data were augmented only with respect to their orientation. The weights of all networks were randomly initialized before training. Five different CNNs architectures were considered for the construction of EPU-CNN models. In detail, two indicative CNN architectures, namely, $Base_I$ and $Base_{II}$, along with VGG$_{16}$, ResNet$_{50}$ and

Table 1. Classification Results (AUC) of EPU-CNN and CNN models

| Models | Datasets | | | | | | |
|---|---|---|---|---|---|---|---|
| | Banapple | KID | Endovis | Kvasir | ISIC 2019 | | |
| | | | | | Ca.vs.Ne. | Ca.vs.Me. | Me.vs.Ne. |
| $EPU_I$ | 0.91 ± 0.01 | 0.94 ± 0.02 | <u>0.97 ± 0.01</u> | 0.87 ± 0.02 | 0.96 ± 0.03 | <u>0.92 ± 0.01</u> | <u>0.94 ± 0.02</u> |
| $EPU_{II}$ | **0.92 ± 0.01** | 0.93 ± 0.01 | **0.97 ± 0.01** | **0.92 ± 0.01** | 0.94 ± 0.02 | 0.91 ± 0.01 | **0.94 ± 0.01** |
| $EPU_{VGG}$ | 0.90 ± 0.01 | 0.93 ± 0.01 | 0.89 ± 0.03 | 0.88 ± 0.01 | **0.97 ± 0.04** | 0.90 ± 0.01 | 0.89 ± 0.03 |
| $EPU_{ResNet}$ | 0.84 ± 0.04 | 0.86 ± 0.06 | 0.84 ± 0.01 | 0.79 ± 0.04 | 0.88 ± 0.08 | 0.78 ± 0.06 | 0.86 ± 0.04 |
| $EPU_{DenseNet}$ | 0.90 ± 0.03 | 0.93 ± 0.05 | 0.87 ± 0.01 | 0.90 ± 0.03 | <u>0.97 ± 0.03</u> | 0.92 ± 0.02 | 0.92 ± 0.03 |
| $Base_I$ | <u>0.92 ± 0.02</u> | <u>0.96 ± 0.02</u> | 0.96 ± 0.01 | <u>0.91 ± 0.01</u> | 0.90 ± 0.02 | **0.94 ± 0.03** | 0.93 ± 0.02 |
| $Base_{II}$ | 0.91 ± 0.01 | **0.97 ± 0.01** | 0.97 ± 0.01 | 0.91 ± 0.01 | 0.93 ± 0.01 | 0.92 ± 0.04 | 0.93 ± 0.04 |
| $VGG_{16}$ | 0.90 ± 0.01 | 0.90 ± 0.04 | 0.93 ± 0.01 | 0.85 ± 0.01 | 0.87 ± 0.01 | 0.90 ± 0.03 | 0.92 ± 0.02 |
| $ResNet_{50}$ | 0.89 ± 0.02 | 0.92 ± 0.03 | 0.88 ± 0.10 | 0.87 ± 0.09 | 0.69 ± 0.12 | 0.90 ± 0.02 | 0.92 ± 0.03 |
| $DenseNet_{169}$ | 0.88 ± 0.04 | 0.94 ± 0.05 | 0.90 ± 0.12 | 0.88 ± 0.01 | 0.76 ± 0.09 | 0.90 ± 0.01 | 0.91 ± 0.03 |
| $TT_{VGG}$ | 0.82 ± 0.04 | 0.91 ± 0.03 | 0.93 ± 0.05 | 0.85 ± 0.04 | 0.88 ± 0.03 | 0.76 ± 0.01 | 0.81 ± 0.02 |

DenseNet$_{169}$ were incorporated as base models in the EPU-CNN framework. These models were selected to demonstrate the generality of the proposed framework, $i.e.$, its applicability to rendering different conventional CNN architectures interpretable. Regarding the architecture of the indicative CNN architectures, $Base_I$, consists of 3 convolutional blocks in total, followed by an FCNN. The first two convolutional blocks are identical and include two convolutional layers followed by a max-pooling and a batch normalization layer. The convolutional layers of these blocks have a depth size of 64 and 128 respectively. The following convolutional block consists of three convolutional layers with a depth size of 256 followed by a max-pooling and a batch normalization. All the kernels of the convolutional layers had a size of 3×3. $Base_{II}$, follows the same architecture with $Base_I$ with an additional convolutional block, in the beginning of the architecture, utilizing an inception module. $Base_I$, $Base_{II}$, VGG$_{16}$, ResNet$_{50}$ and DenseNet$_{169}$ were used for the construction of $EPU_I$, $EPU_{II}$, $EPU_{VGG}$, $EPU_{ResNet}$ and $EPU_{DenseNet}$, respectively. The evaluation followed a 10-fold cross validation procedure with the average Area Under the receiver operating Characteristic (AUC) score among all folds. The AUC was selected as an overall summary measure of binary classification performance, which unlike accuracy, is relatively robust for datasets with imbalanced class distributions [52]. The performance of all models is summarized in Table 1. The best results are in boldface typesetting and the results ranked second are underlined. It can be observed that the results obtained by the EPU-CNN models indicate an overall better or comparable classification performance to their non-interpretable counterparts, $i.e.$, $Base_I$, $Base_{II}$, VGG$_{16}$, ResNet$_{50}$ and DenseNet$_{169}$. In detail, on Banapple, Endovis-MICCAI, Kvasir and ISIC 2019 ($Me.$ $vs.$ $Ne.$)

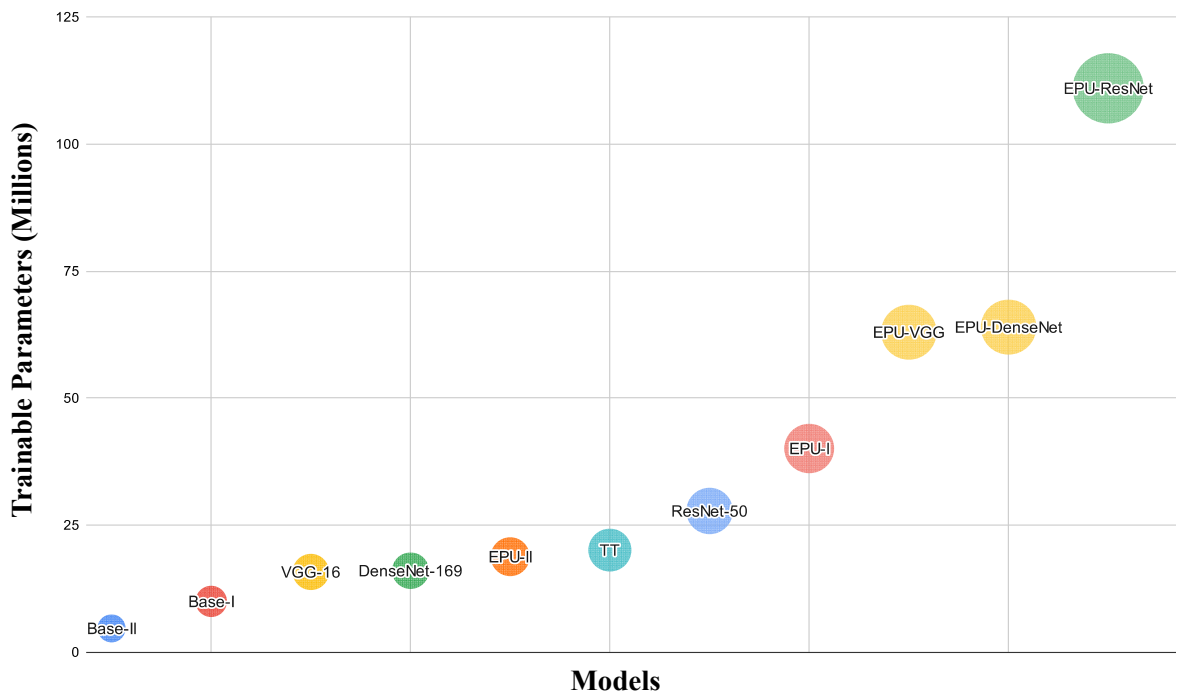

Figure 4. Visualization of the complexity of the compared models in terms of the number of trainable network parameters.

$EPU_{II}$ provided substantially better results when compared to the other EPU-CNN and the majority of base models. Figure 4 illustrates the number of trainable parameters of each model. It can be observed that the complexity of an EPU-CNN model is analogous to that of its base model. Additionally, an EPU-CNN can provide competent results even with base models of low complexity, $i.e.$, $EPU_I$ and $EPU_{II}$ utilize ~40 and ~19 million parameters, respectively;

however, they provide higher classification performance when compared to more complex EPU-CNN models. Furthermore, the less complex $EPU_{II}$ has comparable or even better classification performance when compared to $EPU_I$ while it outperforms substantially ResNet$_{50}$ that is a more computationally demanding base model (~27M parameters). On the other hand, the inherently interpretable (TT) model that has a similar complexity to $EPU_{II}$, $i.e.$, ~20M parameters, provides the lowest overall classification performance amongst all models.

## 3.3 Quantitative Interpretability Analysis

To quantitatively evaluate the interpretability of the proposed framework we exploited the properties of the Banapple benchmark dataset. Banapple is suitable for this purpose because our perception of the class-related objects is directly associated with the way we categorize them, based on their visual attributes regarding color and shape [29]. Therefore, the subsequent task of annotating the images of Banapple did not require any domain-specific knowledge. Images of bananas and apples have distinguishable characteristics with respect to all PFMs utilized by the EPU-CNN models, $i.e.$, *light-dark*, *coarse-fine*, *blue-yellow* and *green-red*. Thus, in the case of a correct class prediction, ideally, all RSSs should trend towards the same direction, as indicated by the sign of an RSS, $e.g.$, all RSSs for an apple image should be positive, whereas for a banana image should be negative. Hence, given that the EPU-CNN models in this study use four PFMs, a ground truth, $\boldsymbol{y}_{int}$, and predicted, $\tilde{\boldsymbol{y}}_{int}$, interpretability label is expressed as follows:

$$\boldsymbol{y}_{int}(y) = \begin{cases} (1,1,1,1), if\ y = 1 \\ -(1,1,1,1), if\ y = 0 \end{cases} \tag{7}$$

$$\tilde{\boldsymbol{y}}_{int}\left(EPU_{CNN}(\boldsymbol{I};\ \eta_{\{n\}})\right) = (sign(\boldsymbol{C}_1(I_1;\ \eta_1)), \dots, sign(\boldsymbol{C}_4(I_4;\ \eta_4))) \tag{8}$$

where $y$ is the ground truth class label of an image $I$, 1 and 0 denotes the apple and banana class respectively whereas $sign(\boldsymbol{C}_i(I_i;\ \eta_i))$ returns the sign of an RSS. Given a set of ground truth and predicted interpretability label pairs, the interpretability accuracy $a_{int}$, is calculated as the average Jaccard Index[53], $J(\cdot)$, among them, as follows:

$$a_{int} = \frac{1}{k}\sum_{j=1}^{k} J\left(\boldsymbol{y}_{int}(y_j), \tilde{\boldsymbol{y}}_{int}\left(EPU_{CNN}(\boldsymbol{I}_j;\ \eta_{\{n\}})\right)\right) \tag{9}$$

Table 2. Interpretability accuracy results of EPU-CNN models.

| Metric | EPU-CNN Models | | | | |
|---|---|---|---|---|---|
| | $EPU_I$ | $EPU_{II}$ | $EPU_{VGG}$ | $EPU_{ResNet}$ | $EPU_{DenseNet}$ |
| $a_{int}$ (%) | $72.40 \pm 1.51$ | $72.62 \pm 1.63$ | $66.64 \pm 2.21$ | $62.90 \pm 4.13$ | $64.62 \pm 2.24$ |

$EPU_I$, $EPU_{II}$ achieved the highest $a_{int}$ with a score of 72.40±1.51% and 72.62±1.63% respectively. This means that the capacity of both $EPU_I$ and $EPU_{II}$ models is comparable with respect to their capacity to interpret the classification of bananas and apples. Since $EPU_{II}$ achieves a better overall classification performance, $a_{int}$ score and it is more computationally efficient, it has been chosen for the qualitative investigation of interpretability that is presented in the following sections.

## 3.4 Ablation Study

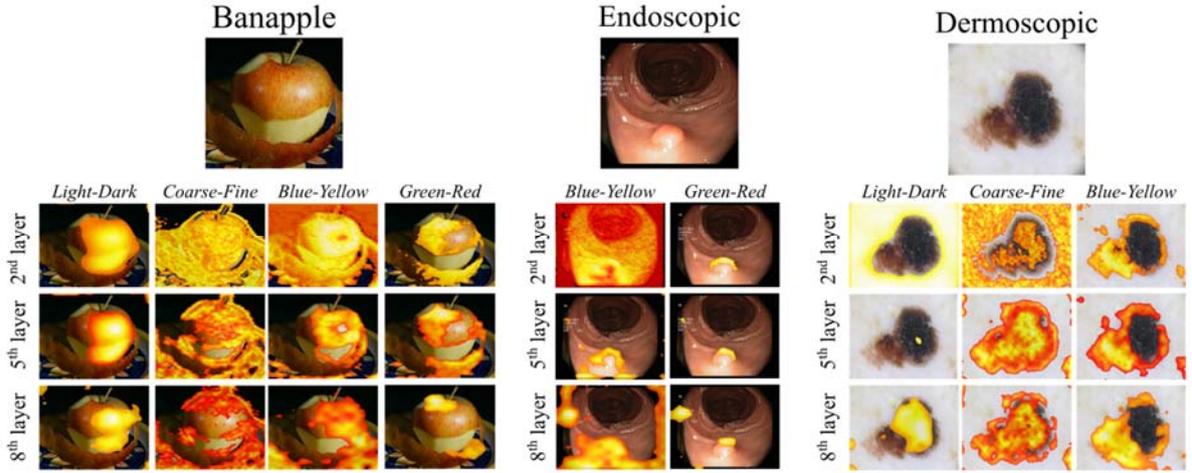

Figure 5. Example of PRMs generated by features maps extracted from different layers of EPU_II.

An ablation study was performed to determine the impact of layer selection to the construction of PRMs, using $EPU_{II}$ as the best performing model. The feature maps estimated by 3 different layers have been chosen for the construction of the respective PRMs. Each of these layers corresponded to the last layer of each convolutional block of $EPU_{II}$.

Figure 5 illustrates indicative PRMs constructed using feature maps estimated by different convolutional layers on predictions from the Banapple, Kvasir and ISIC2019 datasets. As it can be observed, the regions identified as meaningful regarding each PFM are approximately consistent with each other regardless of the degree of abstraction that each set of feature maps encodes. However, the feature maps estimated by the intermediate $5^{th}$ layer provide less noisy PRMs that highlight with more precision the areas on the input image that are estimated to be meaningful with respect to each PFM.

## 3.5 Qualitative Interpretability Analysis

The qualitative analysis of EPU-CNN was investigated by considering both PRMs, global and local bar-charts generated by the $EPU_{II}$ model, for each dataset. Given a validation set of images with *a priori* known class memberships, global bar-charts are constructed by averaging the RSSs per class, as provided by each sub-network of $EPU_{II}$. Global bar-charts enhance the transparency of the model and reveal the overall contribution of PFMs regarding the data discrimination process. In a global bar-chart, PFMs of low or high significance can be identified by their dataset-wide score, which can lead to the selection of a subset of the most informative PFMs, *i.e.,* by pruning or replacing the sub-networks the PFMs of low significance. The respective results obtained per dataset are provided in the next paragraphs.

*Banapple*: The global bar-charts illustrated in Fig. 6(a) indicate that all the perceptual features contribute to the classification of the images. This result is in accordance with our perceptual understanding [29], since apples and bananas are discriminated with respect to all PFMs considered in this study. Figure 7 illustrates examples of local bar-charts along with the respective PRMs of classified images. Specifically, the images presented in Fig. 7(a) were correctly classified by $EPU_{II}$, and this is reflected by the visualization of the RSSs. The PRMs of each sub-network indicate the regions of the input image which resemble the class that each RSS suggests. For instance, in Fig. 7(a)-B the highlighted areas of the PRMs corresponding to *green-red* and *blue-yellow* are overlayed precisely on the class-related object, *i.e.*, the bananas. Interestingly, the difference between the *light-dark* and *coarse-fine* RSSs can be justified by the obscurity of the highlighted regions of $S_{light-dark}$ and $S_{coarse-fine}$, *i.e.*, both PRMs highlight the table.

Figure 7(b) illustrates wrongly classified images. Notably, each of these images have resemblances to the opposite class with respect to color and shape. For example, in Fig. 7(b)-A the perceptual features of *light-dark*, *coarse-fine* and *blue-yellow*, wrongfully guide the prediction towards the banana class (red). This can be justified since the image contains objects that share characteristics that resemble the banana class, *i.e.*, the shape and color of the hands holding the apple. The RSS of *green-red* however, trends towards the apple class (green) with high magnitude, whereas the respective $S_{green-red}$, highlights the apple. Accordingly, the PRMs of *light-dark* and *coarse-fine* focus on the hands explaining the trend of the respective RSSs towards the banana class. Nevertheless, even though $S_{blue-yellow}$ focuses on the apple, the respective RSSs indicate that the image belongs to the banana class. In Fig. 7(b)-B the *light-dark* and *blue-yellow* RSSs trend towards the apple class (green). The direction of these RSSs towards the incorrect class can be justified considering that the color and orientation of the bananas are not representative of their class. Accordingly, $S_{light-dark}$ and $S_{blue-yellow}$ focus only partially on the banana. Similarly, the shape and color from the inside of the apple in Fig. 7(b)-C is unusual for an apple. Hence, the *coarse-fine* and *red-green* RSSs lean towards the opposite direction.

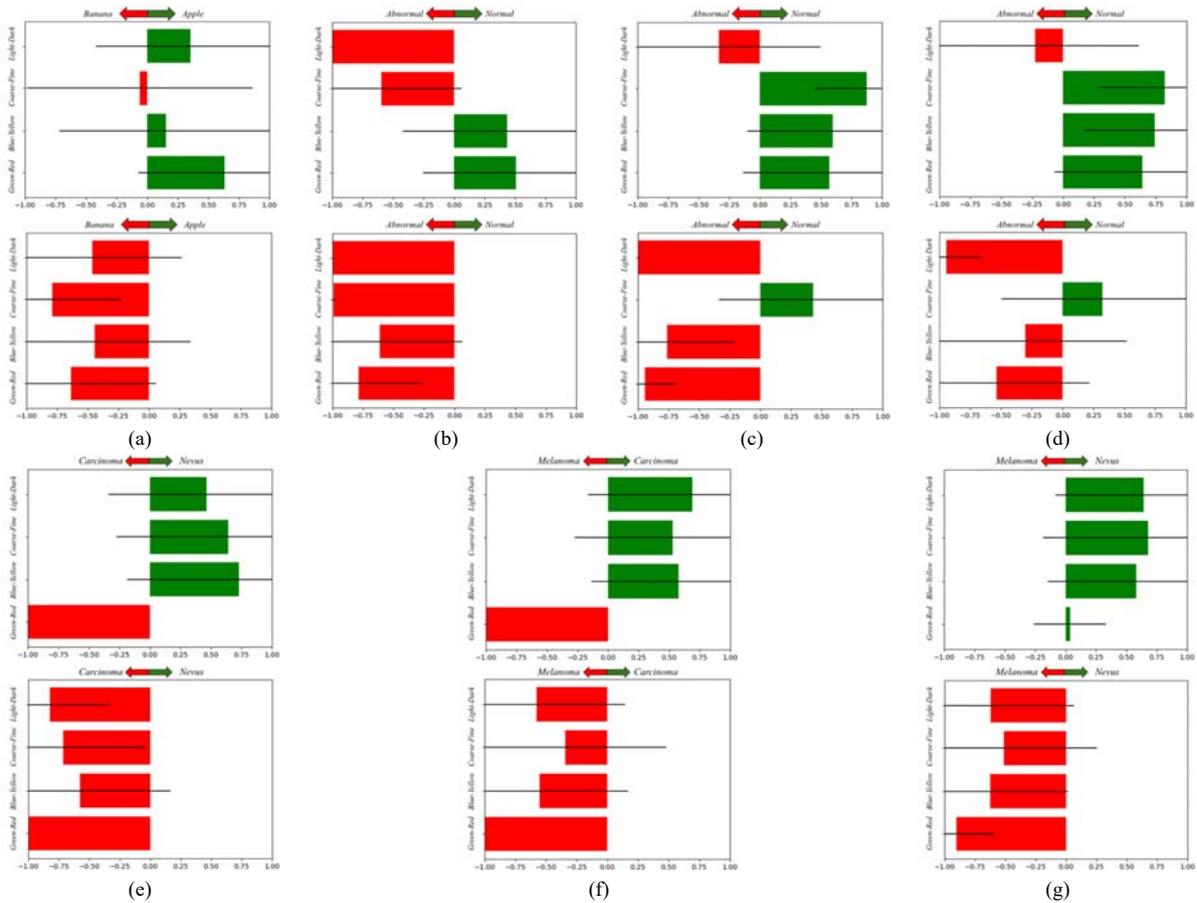

Figure 6. Example of dataset-wide interpretations provided by EPU-CNN on all datasets. Green (positive response) and red (negative response) bars indicating participation the 1 and 0 class respectively, and the black lines indicate the standard deviation. (a) Banapple. (b) KID. (c) MICCAI Endovis 2015. (d) Kvasir. (e-g) ISIC 2019.

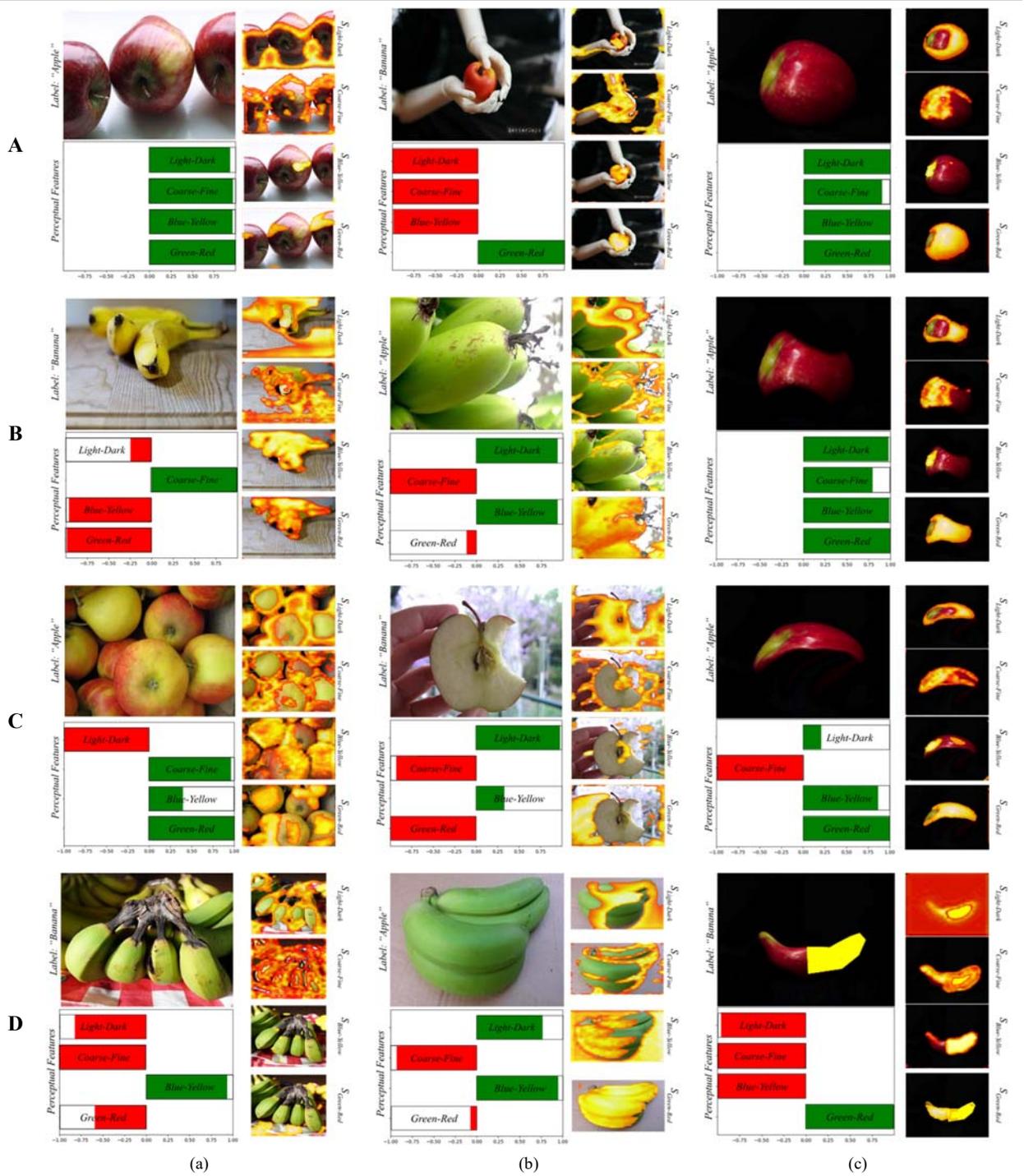

Figure 7. Example of local bar-charts produced by $EPU_{II}$ on images from the Bananapple dataset. The label field indicates the predicted label. (a) Correctly classified images. (b) Wrongly classfied images. (c) Changes in the classification and its interpretation of modified images.

It can be observed that the negative *red-green* and *coarse-fine* scores, have corresponding PRMs that do not focus on the class-related object, *i.e.*, they highlight regions of the hand and the background. Also, the greenish color of the bananas in Fig. 7(b)-D, can be descriptive for both classes (as both bananas and apples can be green), which is also expressed by the disagreement between the relative scores of the color PFMs. Interestingly, the disagreement between the *light-dark* and *coarse-fine* scores can also be justified by the highlighted regions in the respective PRMs, *i.e.,* the outline of the banana object in $S_{coarse-fine}$ and the circular region, resembling an apple in $S_{light-dark}$.

To further investigate the behavior of EPU-CNN, we have chosen an image depicting an apple (Fig. 7(c)-A) which was digitally processed to obtain 3 variations: *a)* to illustrate a bitten apple (Fig. 5(c)-B); *b)* an apple with a shape resembling that of a banana (Fig. 7(c)-C); and *c)* an apple resembling both the shape and color of a banana while maintaining a reddish region (Fig. 7(c)-D). The interpretation changes that can be observed include the following:

a)  When the shape resembles a bitten apple the *coarse-fine* PFM is still guiding the prediction towards the apple class but with greater uncertainty (Fig. 7(c)-B), whereas the $S_{coarse-fine}$ discriminates the image based on its textural variations, *i.e.*, the curvature of the left side of the apple.

b)  When the shape resembles a banana, the *coarse-fine* PFM strongly suggests that the image belongs to the banana class (Fig. 7(c)-C, (c)-D). The magnitude of *light-dark* RSS has also changed, but still trends towards the apple class. The $S_{light-dark}$ and $S_{coarse-fine}$ appear to contribute to the segregation the depicted object; however, only the RSS of *coarse-fine* PFM suggests the opposite class, indicating that is more sensitive to shape variations.

c)  When the yellow region is added, the *light-dark* and *green-yellow* PFMs guide the prediction to the banana class. However, the *green-red* RSS trends towards the apple class. As expected, $S_{blue-yellow}$ and $S_{green-red}$ focus on the yellow and red segments of the object respectively (Fig. 7(c)-D). This justifies the trend of each PFM towards either class, *i.e.*, yellow and red are representative colors of banana and apple class respectively.

These interpretations reveal that the *coarse-fine* PFM enables the respective sub-network to respond to different shape variations and infer relevant decisions. In addition, the color related PFMs, *i.e.*, *blue-yellow* and *green-red*, are very sensitive to the class-related colors and it is clearly reflected both in the respective PRMs and RSSs. When both the class-related colors, *i.e.*, *yellow* and *red*, cooccur in the image, the *blue-yellow* and *green-red* PFM guide the prediction towards the banana and apple class respectively.

*Endoscopic Datasets:* The experiments showed that $EPU_{ll}$ tend to discriminate normal from abnormal images of the endoscopic datasets mainly based on the *blue-yellow* and *green-red* PFMs. This is illustrated in the respective global bar-charts (Fig. 6 (b-d)). As it can be observed, the *light-dark* and *coarse-fine* are biased towards a specific class, in all endoscopic datasets. On the other hand, the chromatic PFMs are the main contributors to the correct classification predictions. This finding is in accordance with the literature since it has been proven that color has a leading role in finding abnormalities in the gastrointestinal tract [37].

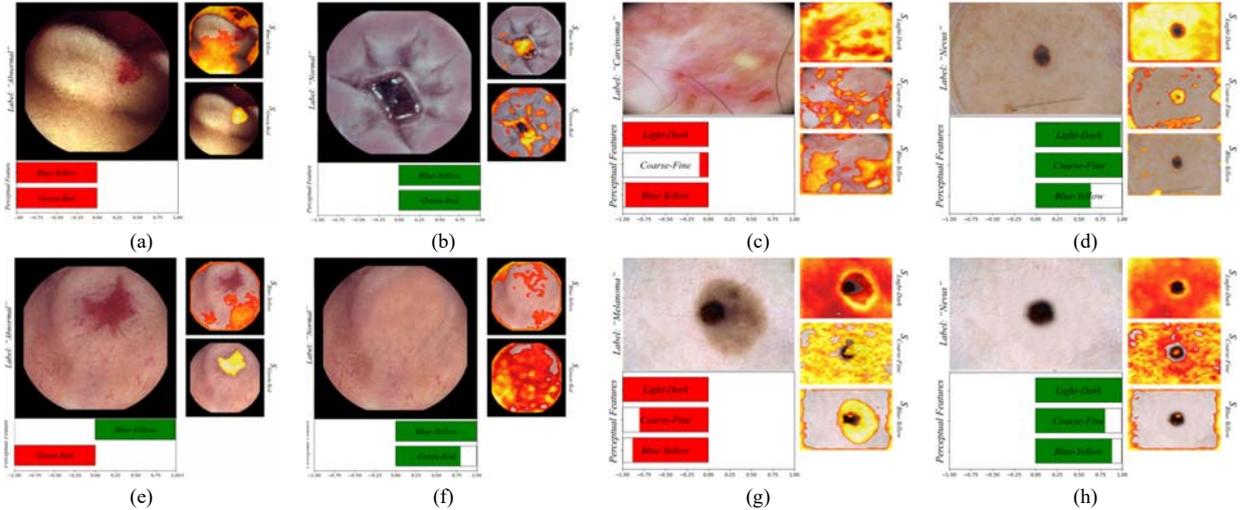

Figure 8. Example of EPU-CNN interpretations, as generated by $EPU_{II}$, on biomedical images. The label field indicates the predicted label. (a) Abnormal and (b) normal endoscopic image; (c) Carcinoma and (d) (normal) nevus skin lesion; (e) Abnormal endoscopic image and (f) modification of (e) to resemble a normal endoscopic image; (g) Melanoma skin lesion and (h) modification of (g) to resemble nevus.

An example of local bar-charts visualizing the prediction interpretations of EPU-CNN on endoscopic images is presented in Fig. 8(a, b). Since the *light-dark* and *coarse-fine* features are not informative, only the color related PFMs were considered. Figures 8(a, b) illustrate correctly classified endoscopic images of both the normal and abnormal class. In the case of the image depicting an abnormality (Fig. 8(a)), the $S_{green-red}$ indicates that the focus of the sub-network that corresponds to the *green-red* PFM focuses on the abnormality, *i.e.*, blood, whereas $S_{blue-yellow}$ focuses on normal tissue and only partially on the abnormal region.

To assess the behavior of $EPU_{II}$ in a more controlled way in the endoscopic datasets, we proceeded to digitally process an endoscopic image and create different conditions for their classification to the normal and the abnormal classes. Indicative examples are presented in Fig. 8 where the abnormal region of Fig. 8(e) is removed, resulting in the synthetic image of Fig. 8(f). The qualitative result of this process, considering only the chromatic PFMs, is reflected in the RSSs and the PRMs of Fig. 8(f). Specifically, it can be noticed that by replacing the abnormal region with normal tissue, the RSS of the *green-red* PFM shifts from trending towards the abnormal class (red) to the normal class (green). Furthermore, the RSS of the *green-red* PFM, in the absence of an abnormality, indicate that the respective subnetwork focuses on normal tissue.

*Dermoscopic Datasets:* The evaluation of the interpretability of $EPU_{II}$ on the dermoscopic datasets revealed that all the PFMs participate actively in the classification process with an exception to *green-red* PFM that appears biased towards either the Melanoma or Carcinoma class on all trials (Fig. 6(e-g)). This is an indication that the PFM of *green-red* is not informative to the network regarding the classification of dermoscopic images. Furthermore, as it can be observed in Fig. 6(e-g) the classification process of $EPU_{II}$ is relying on both chromatic and textural cues (*i.e.*, *blue-yellow, light-dark* and *coarse-fine*) that are also considered by the ABCD rule of skin lesion classification to assess the malignancy of a lesion [54].

An example of local bar-charts of classified dermoscopic images are illustrated in Fig. 8(c, d). The local bar-chart includes the most informative PFMs, *i.e.*, *light-dark*, *coarse-fine* and *blue-yellow*. In Fig. 8(c, d) all RSSs are trending, correctly, towards the abnormal (carcinoma, red) and normal (nevus, green) class respectively. In the case of the carcinoma (Fig. 8(c)), $S_{light-dark}$ focuses on the entirety of the image, whereas $S_{coarse-fine}$ and $S_{blue-yellow}$ focuses on regions with color variations, *e.g.*, on the yellow spot and little cuts on the lest and bottom side of the image respectively. In the case of the nevus (Fig. 8(d)), $S_{light-dark}$ and $S_{coarse-fine}$ isolate the lesion by segregating it from the rest of the image, either by focusing on it or around it, whereas $S_{blue-yellow}$ indicates only a slight attention of the network to the lesion. Similarly, to the other datasets, we proceeded to digitally modify the image of Fig. 8(g) that illustrates a melanoma to resemble a nevus. The modification was implemented according to the rule-based diagnostic criteria expressed by the ABCD rule[54]; in detail, we removed the part of the lesion that introduced color variation on the same mole and obtained a more symmetrical shape. The qualitative results of this process are illustrated in Fig. 8(h), where it is shown that after the modification all the RSSs trend towards the nevus class (green). Furthermore, the $S_{blue-yellow}$ PRM, in the absence of an abnormal region, indicate that the respective subnetwork does not focus on the skin lesion. The $S_{light-dark}$ and $S_{coarse-fine}$ PRMs seem to maintain a similar behavior with the unmodified image.

## 3.6 Comparison with State-Of-The-Art Interpretable Methods

Even though there is an increasing research interest regarding the interpretation of CNNs, there is still not a standard procedure to evaluate and compare the interpretable output. Nevertheless, a qualitative comparison can reveal strengths and weaknesses of such methods. In this study, the interpretations that EPU-CNN provides are qualitatively compared to seven methodologies that have been proposed to interpret CNNs and have been also widely used in the literature. These methods provide saliency maps indicating regions or points on the input image that are estimated to be crucial for a prediction inferred by a CNN.

In detail, six *post-hoc* methodologies, namely, Grad-CAM [55], LIME [8], XRAI [56], Shapley Additive exPlanations (SHAP) [57], Smoothgrad [58] and Vanilla Gradients [59],  as well as one inherently interpretable model [51] (TT) were utilized in this evaluation. The *post-hoc* methodologies were applied on the CNN models that achieved the highest performance on each dataset according to Table 1, whereas TT was trained on each dataset from scratch. All the methods provide interpretations in the form of saliency maps while TT can also provide bounding boxes that specify discreetly the estimated region of interest. These methods were selected since they can render CNN models interpretable without the need for training on datasets specifically annotated for interpretable learning, *e.g.*, with annotation regarding the concepts that are depicted on images.

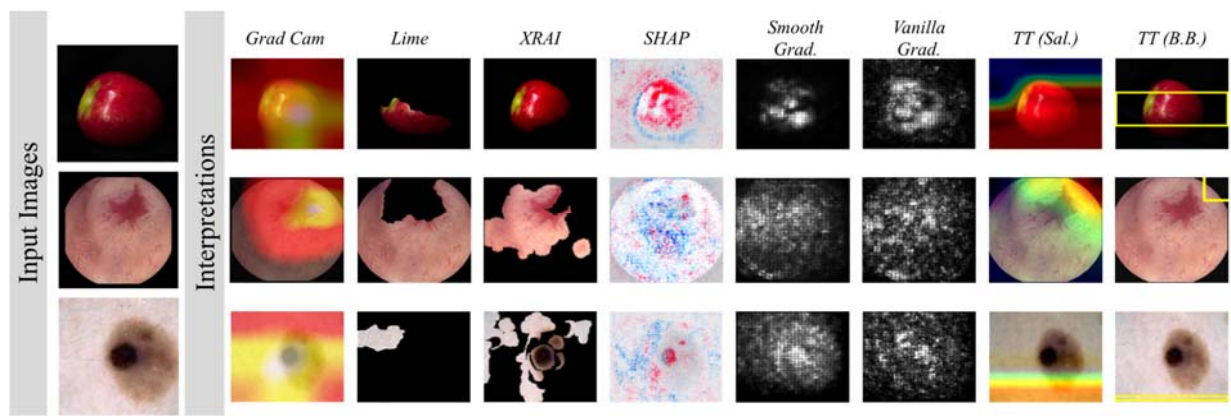

Figure 9. Example of CNN interpretations provided by various methodologies.

Figure 9 summarizes the interpretations provided by each method on exemplary images that are presented in Fig. 7 and 8. All the images have been correctly classified by the respective models that were used. In detail, only XRAI and SHAP were successful at highlighting regions of interest on the images that can be regarded crucial for classification, *i.e.*, areas of the apple, the skin lesion and blood depicted in the endoscopic image. The gradient-based interpretation approaches, *i.e.*, Grad-CAM, Smoothgrad and Vanilla Grad.,  also revealed  that the respective CNN models focus on image regions that can be regarded meaningful; nevertheless, the fuzziness of their visualization makes the communication of their interpretations difficult to comprehend. On the other hand, EPU-CNN can provide different visualizations, that highlight the most relevant regions with respect to each PFM as it was estimated by the layer activations of each subnetwork. This can also be expressed quantitatively since the RSSs indicate the degree to which each highlighted region affects the classification result. Furthermore, the dataset-wide plots that can be constructed by using an EPU-CNN model give insights regarding which PFMs are important for classifying the images of a particular dataset. To the best of our knowledge no other interpretation approach can incorporate all this information to its explanations and simultaneously be applied on non-specialized datasets, *e.g.*, datasets where each image is annotated only with respect to their class membership.

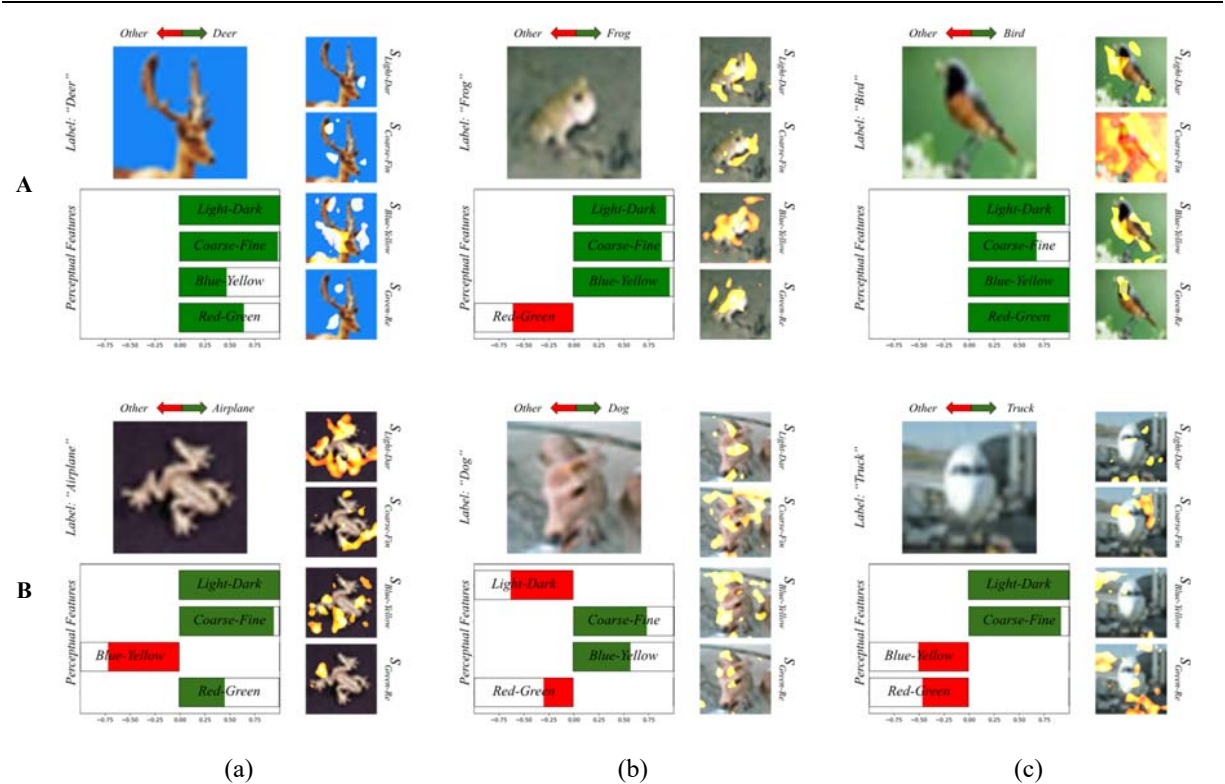

Figure 10. Example of EPU-CNN interpretations, as generated by $EPU_{II}$, on images of the CIFAR-10 dataset. The label field indicates the predicted label. Row A and B illustrate interpretations of correct and wrong prediction, respectively.

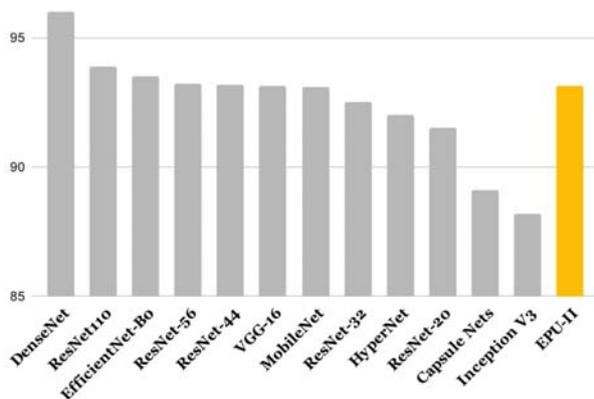

Figure 11. Classification performance in terms of accuracy on the CIFAR-10 dataset.

To demonstrate the applicability of EPU-CNN framework, the best EPU-CNN model in terms of classification performance on the previous experiments, *i.e.*, $EPU_{II}$, was trained and tested on the CIFAR-10 dataset. To train $EPU_{II}$ in a multiclass setting the SoftMax activation function replaced the final log-sigmoid $\sigma(\cdot)$ (Eq. 5) function. Figure 10 illustrates interpretations provided by $EPU_{II}$ on predictions of images on the CIFAR-10 dataset. For consistency with the interpretations of the binary settings, the respective illustrations depict how each PFM drives a prediction towards either the predicted (green) or any other class (red). The rows A and B of Fig. 10 illustrate interpretations of correct and wrong classifications on images included in the CIFAR-10 dataset, respectively. As it can be observed the PRMs generated by the $EPU_{II}$ on the interpretations presented in Fig. 10-A highlight the object of interest with more precision when compared to the wrongly classified images (Fig. 10-B). For example, in Fig. 10(b)–A all the PRMs highlight regions of the frog, and the image has been correctly classified. On the other hand, in Fig. 10(a)-B the PRMs mainly highlight regions around the frog and the respective image is misclassified to the airplane class, based on all the PFMs except from the *blue-yellow*. Furthermore, it can be noticed that the respective RSSs behave similarly, *i.e.*, in Fig. 10(c)-A the PRM of *coarse-fine* highlights the whole image and it does not focus solely on the bird. Accordingly, the respective RSS of *coarse-fine* has a smaller magnitude towards the correct class than the rest of RSSs, which, based on the respective PRMs consider mainly the region of the bird. Moreover, in Fig. 10(b)-B the image that belongs to the deer class is misclassified as a dog. As it can be noticed, the PRMs of *coarse-fine* and *blue-yellow* are mainly focusing on the head region of the deer and the respective RSSs drives the prediction towards the dog class. This can be attributed to the fact that the particular deer does not seem to have horns, and it has a color pattern that matches that of the dog class. This can be further substantiated by observing the image of Fig.10(a)-A. This image depicts a deer that has been correctly classified by the EPU-CNN model. As it can be noticed, the PRMs of *coarse-fine* and *blue-yellow* highlight the head region of the deer where the horns are present. Finally, the image of Fig. 10(c)-B that depicts an airplane is classified to the truck class based on the PFMs of *light-dark* and *coarse-fine*. As it can be noticed, all the PRMs focus on the body of the airplane, and on the wheels, whereas no PRM focuses on the wings. Therefore, the respective RSSs of *light-dark* and *coarse-fine* drive the prediction with a higher magnitude towards the truck class. Figure 11 presents a comparison in terms of classification accuracy among $EPU_{II}$ (orange bar) and other state-of-the-art CNN models [50,49,48,60,61] (gray bars) on the CIFAR-10 dataset. $EPU_{II}$ achieved an accuracy score of 93.31% which is comparable or better than the other models considered. However, a major advantage over the other models is that the EPU-CNN model can provide interpretations regarding the classification outcome.

## 4. Conclusions and Future Work

In this study we propose a novel, generalized framework, called EPU-CNN, that provides a guideline for the development of interpretable CNN models, inspired by GAMs. A model, designed according to EPU-CNN framework, consists of an ensemble of sub-networks with a base CNN architecture that is trained as one. The proposed framework can be used to render conventional CNN model interpretable, by using it as a base model. Each sub-network receives as input a different PFM of an input image, chosen according to the literature of cognitive science and human perception. EPU-CNN is designed in such way enabling human-friendly interpretations of its classification results based on the utilized perceptual features. The interpretations provided EPU-CNN are in the form of RSSs that quantify the resemblance of a perceptual feature to a respective class. These interpretations are complemented by PRMs indicating the image regions where the network focuses to infer its interpretable decisions. Furthermore, EPU-CNN provides spatial expression of an explanation on the input image. Thus said, the most important conclusions of this study can be summarized as follows:

- EPU-CNN models satisfy the need for interpretable models based on human perception, *i.e.*, the proposed framework is able to provide interpretations in accordance with human perception and cognitive science, *e.g.*, EPU-CNN classifies endoscopic images based on the chromatic perceptual features.

- Unlike other inherently interpretable CNN methodologies [20], [28], the classification performance of EPU-CNN models is not affected by their capacity to provide interpretations. In fact, the results obtained from the comparison of EPU-CNN models with respective non-interpretable CNN models, show that their performance is better or at least comparable to that of the non-interpretable models.

- When an image is modified with respect to a perceptual feature, *e.g.*, color, the interpretations derived from the EPU-CNN model change accordingly both on natural and biomedical images (Figs. 6, 7).

- Since EPU-CNN is a generalized framework, it provides a template for the development of interpretable CNNs that fulfill the requirements imposed by current legislations regarding the commercial applicability of ML models.

An aspect that can be considered as a limitation of the proposed framework, is the manual selection of PFMs. The process of the selection, however, enables the user to leverage specific PFMs that are relevant to a particular application and as a result to acquire meaningful interpretations and insights regarding the internal process of an EPU-CNN model. For example, in the case of endoscopic images, the EPU-CNN models considered only the PFMs of color as more important which is in accordance with the respective literature. Most inherently interpretable models that have been proposed in the literature, can only be applied on datasets that are further annotated with respect to human-understandable concepts illustrated in each image, which results in limitation regarding their applicability [18, 19]. The PFM selection of an EPU-CNN model can be considered as a less demanding and time-consuming procedure when compared to the annotation of huge datasets with the human-understandable concepts. Additionally, since the selection of the textural and color perceptual features, that are based on the 2D DWT and *Lab*, respectively, is empirical, as a future work we intend to automate the PFM selection towards a direction that minimizes human intervention and is more compatible with the principles of deep learning.

**Acknowledgements**

We acknowledge support of this work by the project "Smart Tourist" (MIS 5047243) which is implemented under the Action "Reinforcement of the Research and Innovation Infrastructure", funded by the Operational Programme "Competitiveness, Entrepreneurship and Innovation" (NSRF 2014-2020) and co-financed by Greece and the European Union (European Regional Development Fund).